\documentclass[11pt,a4paper]{article}
\usepackage[hyperref]{acl2021}
\usepackage{times}
\usepackage{latexsym}

\usepackage{microtype}

\aclfinalcopy 
\def\aclpaperid{1} 

\usepackage{covington}

\title{Is Sluice Resolution really just Question Answering? \\(Extended Abstract)}

\author{ Peratham Wiriyathammabhum \\
   \texttt{peratham.bkk@gmail.com} \\}

\date{}

\begin{document}
\maketitle
\begin{abstract}
Sluice resolution is a problem where a system needs to output the corresponding antecedents of wh-ellipses. The antecedents are elided contents behind the wh-words but are implicitly referred to using contexts. Previous work frames sluice resolution as question answering where this setting outperforms all its preceding works by large margins. Ellipsis and questions are referentially dependent expressions (anaphoras) and retrieving the corresponding antecedents are like answering questions to output pieces of clarifying information. However, the task is not fully solved. Therefore, we want to further investigate what makes sluice resolution \textit{differ} to question answering and fill in the error gaps. We also present some results using recent state-of-the-art question answering systems which improve the previous work (86.01 to 90.39 F1).
\end{abstract}

\section{Introduction}
Sluicing \cite{ross1969guess} is a variety of ellipsis which uses interrogative words (wh-words) to elide everything but themselves subject to the antecedents. The preceding antecedents and the elliptical content have various syntactic and semantic mismatches. The task is to retrieve the elided content from context (including the antecedents). 
\begin{examples}
\item He never spoke this way during his first two camps and regular seasons. Why not?
\end{examples}
We can see that the first sentence is elided in the second sentence and the remaining is only `Why not?'.

\section{Related Work}
\cite{anand2015annotating} introduced the sluice resolution task along with a corpus of 3,103 examples (ESC) based on the New York Times subset of the Gigaword Corpus \cite{graff2003english}. Their subsequent version \cite{anand2021sluicing} contains 4,700 examples.  

\cite{anand2016antecedent} is the first sluice resolution system that constructs fifteen linguistically sophisticated features in syntax and discourse. Then, a linear combination (a maxent model) is learned via a simple hill-climbing procedure. This set a preliminary result of 0.67 F1 score on ESC. \cite{ronning2018sluice} argued that syntactic features do not generalize to other languages and domains. Therefore, a multi-task learning system was proposed to learn auxiliary syntactic features without a partial parser. This boosts the F1 score from 0.67 to 0.7 for the ESC dataset. \cite{anand2021sluicing} reframed ellipsis resolution problems as machine comprehension tasks that can be solved by various off-the-shelf extractive question answering systems (QA). Interestingly, this gives a huge performance boost to 0.86 F1 score for sluice resolution on the ESC dataset using a vanilla BERT model \cite{devlin2019bert}. 

Recently, there are so many proposed models for QA, mostly based on BERT. We consider SpanBERT \cite{joshi2020spanbert} and LUKE \cite{yamada2020luke}. They are state-of-the-art systems for SQuAD v1.1 \cite{rajpurkar2016squad}. SpanBERT was tailored for text span prediction problems by changing the masking scheme of BERT from masking tokens to masking spans along with a novel span-boundary objective function (SBO) to predict the content of the whole span at its boundary. The intuition is language meaning should be better represented as relations between spans rather than only between tokens \cite{46490}. LUKE is the current state-of-the-art on SQuAD v1.1 which utilizes knowledge-based embeddings based on Wikipedia so this provides prior external knowledge about entities. LUKE also extends RoBERTa \cite{liu2019roberta} to the entity-aware self-attention mechanism which supports both word and entity vector inputs. 

\section{Experiments}
\subsection{Recent Models on Question Answering Formulation}
Following the recent state-of-the-art in sluicing resolution \cite{anand2021sluicing}, we reframe sluicing resolution as question answering (QA) and employ recent state-of-the-art QA models. We consider SpanBERT \cite{joshi2020spanbert}. We use the ESC dataset along with the splitting from \cite{ronning2018sluice} and we convert them to the SQuAD v1.1 format \cite{rajpurkar2016squad} as in \cite{anand2021sluicing} so the results are directly comparable. The splitting consists of \textit{1.4k} training examples, \textit{480} validation examples and \textit{992} test examples. Following the single-task setup, we train single models on the training set and test on the separated test set as in the standard evaluation setting. 

\begin{table}[] 
\centering
\caption{Sluice resolution as question answering: Sluice resolution test scores in token-level F1 for single models on ESC.}
\label{table:exp1}
\begin{tabular}{l c}
\hline
Model    & F1  \\
\hline
BERT in \cite{anand2021sluicing} & 85.10 \\
\hline
SpanBERT \cite{joshi2020spanbert} &  \textbf{88.18}  \\
LUKE \cite{yamada2020luke} & \underline{\textbf{90.39}} \\
\hline
\end{tabular}
\vspace{-3\baselineskip}
\end{table}

From Table \ref{table:exp1}, recent models outperform BERT as expected. We only use BERT-base (or RoBERTa-base) for all models. SpanBERT has an \textit{3.08} F1 score improvement while LUKE has an \textit{5.29} F1 score improvement over training BERT on the ESC training set. Both SpanBERT and LUKE outperform the previous best result which is training BERT on a concatenation of ESC and SQuAD v1.1 by \textit{2.17} and \textit{4.28} F1 score respectively. The results suggest that recent advancements in question answering also apply to sluice resolution in some way. 

\subsection{Transferring Knowledge from other tasks via Fine-tuning}
The ESC dataset is relatively small (few thousand examples) compared to other recent datasets such as SQuAD or OntoNotes (tens to hundreds thousands of examples). Therefore, it might be helpful to use the fine-tuning scheme (FT) so the knowledge from another task might help sluice resolution as in the multi-task model from \cite{ronning2018sluice}. We pre-train SpanBERT and LUKE on the SQuAD v1.1 dataset then finetune on ESC. We reduce the learning rate by half after every epoch during fine-tuning.

We found that the random seed affects the improvements by a lot in this setting for SpanBERT. SpanBERT performance improves consistently but the best results vary from \textit{88.5} to \textit{89.5} with different seeds. We conclude that the fine-tuning scheme can increase the F1 score but it might be from the optimization perspective of some latent underfitting issue when training only on ESC.  \cite{anand2021sluicing} reframes sluice resolution as question answering but they did not consider the pre-training/fine-tuning scheme into account. They considered combining datasets and training the models on various combinations instead where they found the combination of SQuAD and ESC make the most improvement. We also consider this setting and found that the result is worse than fine-tuning for SpanBERT. For LUKE, both fine-tuning and using the concatenation of ESC and SQuAD do not improve the performance. We hypothesize that the knowledge-based embedding from Wikipedia already provides necessary information for sluice resolution which might worsen when being pre-trained on question answering. This also means that entity knowledge matters a lot in sluice resolution and is complimentary to just question answering. 

\begin{table}[] 
\centering
\caption{Pre-training/Fine-tuning QA models for sluice resolution: Test scores in token-level F1 on ESC.}
\label{table:exp2}
\begin{tabular}{l c c}
\hline
Model  & concat  &  fine-tuning  \\
 & data &   \\
\hline
BERT in  & 86.01 &  -\\
\cite{anand2021sluicing} & & \\
\hline
SpanBERT  &  \textbf{88.55}  & \textbf{89.51}\\
\cite{joshi2020spanbert} &  & \\
LUKE  & \textbf{89.95} & \textbf{89.64}  \\
\cite{yamada2020luke} &  & \\
\hline
\end{tabular}
\vspace{-3\baselineskip}
\end{table}

\section{Summary}
We consider sluice resolution which is finding the elided information for wh-ellipses. Recent results suggest posing this task as question answering as well as for the trend in many other tasks \cite{gardner2019question}. We consider some recent state-of-the-art question answering systems and hypothesize that entity knowledge, for example, might be helpful in addition to just posing the problem as question answering. Also, the task may not really be question answering since transferring knowledge from question answering datasets seems to be not so helpful.

\bibliographystyle{acl_natbib}
\bibliography{paper_bib}


\end{document}